\documentclass[letterpaper,journal]{IEEEtran}
\usepackage[utf8]{inputenc}
\usepackage{url}
\usepackage{cite}
\usepackage{amsmath,amssymb,amsfonts}
\usepackage{textcomp}
\usepackage{lettrine}
\usepackage{caption}
\usepackage{subcaption}
\usepackage[export]{adjustbox}
\newcommand{\subparagraph}{}
\usepackage{leftidx}
\usepackage{siunitx}
\usepackage{multirow}
\usepackage{balance}
\usepackage{booktabs}

\usepackage{hyperref}
\usepackage{algorithm2e}
\RestyleAlgo{ruled}
\usepackage{mathtools}
\usepackage{setspace}

\begin{document}

\title{\LARGE \bf \textit{S-Graphs+}: Real-time Localization and Mapping leveraging Hierarchical Representations 
}

\author{Hriday Bavle$^{1}$, Jose Luis Sanchez-Lopez$^{1}$, Muhammad Shaheer$^{1}$, \\ Javier Civera$^{2}$ and Holger Voos$^{1}$ 
\thanks{*This work was partially funded by the Fonds National de la Recherche of Luxembourg (FNR), under the projects C19/IS/13713801/5G-Sky, by a partnership between the Interdisciplinary Center for Security Reliability and Trust (SnT) of the University of Luxembourg and Stugalux Construction S.A., by the Spanish Government under Grant PID2021-127685NB-I00 and by the Arag{\'o}n Government under Grant DGA T45 17R/FSE.
For the purpose of Open Access, the author has applied a CC BY public copyright license to any Author Accepted Manuscript version arising from this submission.}
\thanks{$^{1}$Authors are with the Automation and Robotics Research Group, Interdisciplinary Centre for Security, Reliability and Trust, University of Luxembourg. Holger Voos is also associated with the Faculty of Science, Technology and Medicine, University of Luxembourg, Luxembourg.
\tt{\small{\{hriday.bavle, joseluis.sanchezlopez, muhammad.shaheer, holger.voos\}}@uni.lu}}%
\thanks{$^{2}$Author is with I3A, Universidad de Zaragoza, Spain
{\tt\small jcivera@unizar.es}}%
}

\maketitle

\begin{abstract}

In this paper, we present an evolved version of Situational Graphs, which jointly models in a single optimizable factor graph (1) a pose graph, as a set of robot keyframes comprising associated measurements and robot poses, and (2) a 3D scene graph, as a high-level representation of the environment that encodes its different geometric elements with semantic attributes and the relational information between them.

Specifically, our \textit{S-Graphs+} is a novel four-layered factor graph that includes:
(1) a keyframes layer with robot pose estimates,
(2) a walls layer representing wall surfaces,
(3) a rooms layer  encompassing sets of wall planes,
and (4) a floors layer gathering the rooms within a given floor level. 
The above graph is optimized in real-time to obtain a robust and accurate estimate of the robot's pose and its map, simultaneously constructing and leveraging high-level information of the environment. To extract this high-level information, we present novel room and floor segmentation algorithms utilizing the mapped wall planes and free-space clusters.

We tested \textit{S-Graphs+} on multiple datasets, including simulated and real data of indoor environments from varying construction sites, and on a real public dataset of several indoor office areas.  
On average over our datasets, \textit{S-Graphs+} outperforms the accuracy of the second-best method by a margin of $10.67\%$, while extending the robot situational awareness by a richer scene model. Moreover, we make the software available as a docker file.

\noindent \textbf{Project web:} \url{https://snt-arg.github.io/s_graphs_docker/}
\end{abstract}


\section{Introduction}

\IEEEPARstart{R}{obots} require a deep understanding of the situation for their autonomous and intelligent operations \cite{sa_survey}.
Works like \cite{3d_scene_graph}, \cite{dynamic_scene_graph}, \cite{scene_graph_fusion}, \cite{hydra} generate 3D scene graphs modeling the environment with high-level semantic abstractions (such as chairs, tables, or walls) and their relationships (such as a set of walls forming a room or a corridor). While providing a rich understanding of the scene, they typically rely on separate SLAM methods, such as \cite{loam}, \cite{floam}, \cite{lego-loam}, that previously estimate the robot's pose and its map using metric/semantic representations without exploiting this hierarchical high-level information of the environment. Methods like \cite{hydra} do optimize the full 3D scene graphs but only after detection of appropriate loop closures. Thus, in general, 3D scene graphs are not tightly and continuously optimized in a factor graph.      

\begin{figure}[t]
    \centering
    \includegraphics[width=0.43\textwidth]{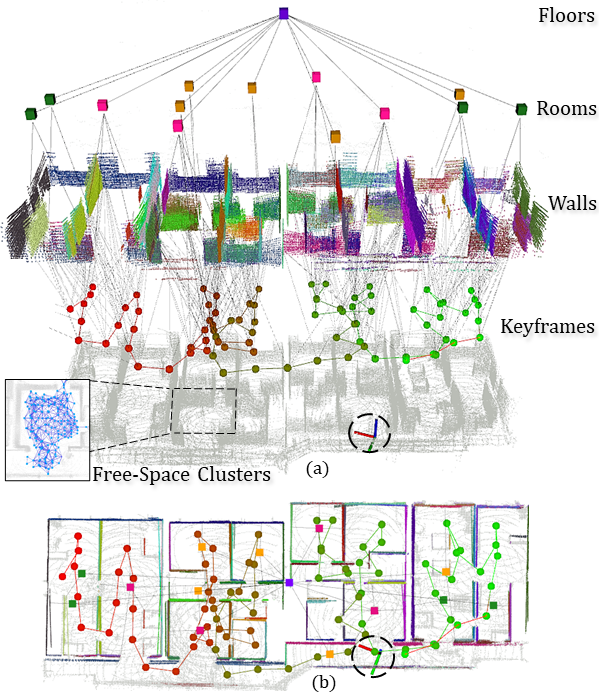}
    \caption{\textit{S-Graph+} built using a legged robot (circled in black) as it navigates a real construction site consisting of four adjacent houses. (a) 3D view of the four-layered hierarchical optimizable graph. The zoomed-in image shows a partial view of the free-space clusters utilized for room segmentation. (b) Top view of the graph.}  
    \label{fig:scene_graph}
\end{figure} 

Our previous work \textit{S-Graphs} \cite{s_graphs} proposed for the first time a tight coupling of geometric LiDAR SLAM with 3D scene graphs in a single optimizable factor graph, demonstrating state-of-the-art metrics. However, it came with multiple limitations that we overcome in this work with our new \textit{S-Graphs+}, with updated front-end and back-end relying on 3D LiDAR measurements.

Our new \textbf{front-end} (Sec.~\ref{proposed_method}) contributes over \textit{S-Graphs} with (1) a novel \textit{room segmentation} algorithm using free-space clusters and wall planes, providing higher detection recall and removing most heuristics of the \textit{S-Graphs} counterpart;
(2) an additional \textit{floor segmentation} algorithm extracting the floor centers using all the currently extracted wall planes.

The new \textbf{back-end} (Sec.~\ref{sec:back_end}) consists of an improved real-time optimizable factor graph composed of four layers. 
A \textit{keyframes layer} constraining a sub-set of robot poses at specific distance-time intervals. 
A \textit{walls layer} constraining of the wall plane parameters and linked to the keyframes using pose-plane constraints. Both the layers are analogous to \textit{S-Graphs}.  
A \textit{rooms layer} modeling detected rooms to their corresponding wall planes constraining them in a single tightly coupled factor, rather than loosely coupled factors in \textit{S-Graphs}. A \textit{floors layer}, denoting the current floor level in the graph and constraining the rooms at that level, not present in \textit{S-Graphs}. See Fig.~\ref{fig:scene_graph} for an illustrative example of an \textit{S-Graph+} of a real building.


Our main contributions are, therefore, summarized as: 
\begin{itemize}
    \item A novel real-time factor graph organized in four hierarchical layers and a specifically novel room-to-wall factor.  
    \item A real-time extraction of high-level information, specifically novel room and floor segmentation algorithms.
    \item A thorough experimental evaluation in different simulated and real construction/office environments as well as software release for the research community.  
\end{itemize}

\section{Related Works}
\label{sec:related_works}


\subsection{SLAM and Scene Graphs}

The literature on LiDAR SLAM is huge, and there are several well-known geometric approaches like LOAM \cite{loam} and its variants \cite{floam}, \cite{mloam}, \cite{hdl_graph_slam}, and also semantic ones like LeGO-LOAM \cite{lego-loam}, SegMap \cite{segmap}, SUMA++ \cite{suma++} that provide robust and accurate localization and 3D maps of the environments. While geometric SLAM lacks meaning in the representation of the environments, causing failures in aliased environments and limitations for high-level tasks or human-robot interaction, its semantic SLAM counterparts lack in most occasions geometric accuracy and robustness, due to wrong matches between the semantic elements and the limited relational constraints between them.       

Scene graphs, on the other hand, model scenes as structured representations, specifically in the form of a graph comprising objects, their attributes, and the inter-relationships among them. This high-level representation has the potential to boost several relevant challenges in SLAM, such as map compacity or understanding. Focusing on 3D scene graphs for understanding, the pioneering work \cite{3d_scene_graph} creates an offline semi-autonomous framework using object detections from RGB images, generating a multi-layered hierarchical representation of the environment and its components, divided mainly into layers of camera, objects, rooms, and building. \cite{3-d_scene_graph} presents a framework for generating a 3D scene graph using a sequence of images to verify its applicability to visual questioning and answering and to task planning. 3D SSG (Semantic Scene Graph) \cite{3d_ssg} presents a learning method based on PointNet and Graph Convolutions Networks (GCN) to semi-automatically generate graphs for 3D scenes. SceneGraphFusion \cite{scene_graph_fusion} on the other hand, generates a real-time incremental 3D scene graph using RGB-D sequences, accurately handling partial and missed semantic data. 3D DSG (Dynamic Scene Graph) \cite{dynamic_scene_graph} extend the 3D scene graph concept to environments with static parts and dynamic agents in an offline manner, while Hydra \cite{hydra}, presents research in the direction of real-time 3D scene graph generation as well as its optimization using loop closure constraints. 
Though promising in terms of scene representation and higher-level understanding, a major drawback of these models is that they do not tightly couple the estimate of the scene graph with the SLAM state, in order to simultaneously optimize them. They thus in general generate a scene graph and a SLAM graph in an independent manner. Our previous work \textit{S-Graphs} \cite{s_graphs} bridged this gap showcasing the potential of tightly coupling SLAM graphs and scene graphs. However, for several reasons, it was limited to simple structured environments. Our current work \textit{S-Graphs+} overcomes these limitations generating a four-layered hierarchical optimizable graph while simultaneously representing the environment as a 3D scene graph, able to provide an excellent performance even in complex environments.

\subsection{Room Segmentation}

For a robot to understand structured indoor environments, it is necessary to first understand their basic components, such as walls, and their composition into higher-level structures such as rooms. Hence, room identification and segmentation is one of the critical tasks in \textit{S-Graphs+}. In the literature, different room segmentation techniques are presented over pre-generated maps using 2D LiDARs \cite{roomsegmentationreview, maoris, mapsegnet}. Their performance is, however, degraded in presence of clutter. While \cite{rose2} presents a room segmentation approach based on pre-generated 2D occupancy maps in cluttered indoor environments, it still lacks real-time capabilities. Methods such as \cite{3D_semantic_parsing, automatic_room_segmentation, 3d_room_recontruction} perform segmentation of indoor spaces into meaningful rooms, although they require a pre-generated 3D map of the environment and cannot segment it in real-time. Authors in \cite{hydra} present a real-time room segmentation approach to classifying different places into rooms but compared to our approach they do not utilize the walls in the environment to efficiently represent the rooms. Given the current state-of-the-art for room segmentation, there was a need to develop a room segmentation algorithm utilizing wall entities while capable of running in real-time as the robot explores its environment, to simultaneously incorporate this high-level information into the optimizable \textit{S-Graphs+}. 

\section{Overview}
\label{overview}

\begin{figure}[h]
    \centering
    \includegraphics[width=0.45\textwidth]{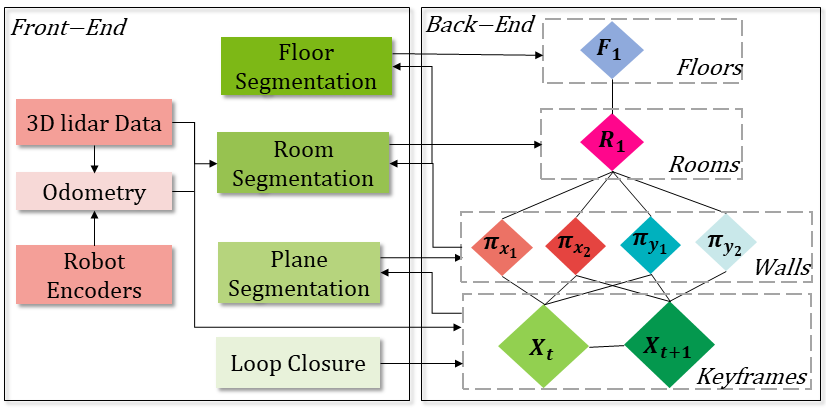}
    \caption{\textit{S-Graphs+} overview. Our inputs are the 3D LiDAR measurements and robot odometry, which are pre-filtered and processed in the front-end to extract wall planes, rooms, floor, and loop closures. Note the four-layered \textit{S-Graph+}, whose parameters are jointly optimized in the back-end.}
    \label{fig:system_architecture}
    \vspace{-2mm}
\end{figure}

The architecture of \textit{S-Graphs+} is illustrated in Fig.~\ref{fig:system_architecture}. Its pipeline can be divided into six modules, and its estimates are referred to four frames: the LiDAR frame $L_t$, the robot frame $R_t$, the odometry frame $O$, and the map frame $M$. $L_t$ and $R_t$ are rigidly attached to the robot and depend on the time instant $t$, while $O$ and $M$ are fixed. The first module receives the 3D LiDAR point cloud in frame $L_t$, which is pre-filtered and downsampled. The second module estimates the robot odometry in frame $O$ either from LiDAR measurements or the robot encoders. \textit{S-Graphs+} is agnostic to the source of odometry, thus it can utilize odometry estimated either from sensor measurements like 3D LiDAR or directly generated from encoders of the robotic platforms. 
Four additional front-end modules generate the four-layered topological graph modeling the understanding of the environment, namely: 1) The plane segmentation module, segmenting and initializing wall planes in the map frame $M$ using the point clouds at each keyframe. 2) The room segmentation module, generating first free-space clusters from the robot poses and 3D LiDAR measurements, and then using such clusters along with the mapped planes to detect room centers in frame $M$. 3) The floor segmentation module, utilizes the information of all the walls in the map to extract the center of the current floor level in frame $M$. 4) Finally, the loop closure module as in \cite{s_graphs}, which utilizes a scan-matching algorithm to recognize revisited places and correct the drift.  

We define the global state as:
\begin{multline}
\mathbf{s} = [\leftidx{^M}{\mathbf{x}}_{R_1}, \ \hdots, \ \leftidx{^M}{\mathbf{x}}_{R_T}, \ \leftidx{^M}{\boldsymbol{\pi}}_{1}, \ \hdots, \ \leftidx{^M}{\boldsymbol{\pi}}_{P}, \\
\leftidx{^M}{\boldsymbol{\rho}}_{1}, \ \hdots, \ \leftidx{^M}{\boldsymbol{\rho}}_{S}, \ \leftidx{^M}{\boldsymbol{\kappa}}_{1}, \ \hdots, \  \leftidx{^M}{\boldsymbol{\kappa}}_{K}, \ \\  \leftidx{^M}{\boldsymbol{\xi}}_{1}, \ \hdots, \ \leftidx{^M}{\boldsymbol{\xi}}_{F}, \ \leftidx{^M}{\mathbf{x}}_{O}]^\top,
\end{multline}

\noindent where $\leftidx{^M}{\mathbf{x}}_{R_t} \in SE(3), \ t \in \{1, \hdots, T\}$ are the robot poses at $T$ selected keyframes, $\leftidx{^M}{\boldsymbol{\pi}}_{i} \in \mathbb{R}^3, \ i \in \{1, \hdots, P\}$ are the plane parameters of the $P$ wall planes in the scene, $\leftidx{^M}{\boldsymbol{\rho}}_{j} \in \mathbb{R}^2, \ j \in \{1, \hdots, S\}$ contains the parameters of the $S$ four-wall rooms and $\leftidx{^M}{\boldsymbol{\kappa}}_{k} \in \mathbb{R}^2, \ k \in \{1, \hdots, K\}$ the parameters of the $K$ two-wall rooms, $\leftidx{^M}{\boldsymbol{\xi}}_{f} \in \mathbb{R}^2, f \in \{1, \hdots \, F \}$ are the $F$ floors levels, and $\leftidx{^M}{\mathbf{x}}_{O}$ models the drift between the odometry frame $O$ and the map frame $M$. 

\section{Front-End} 
\label{proposed_method}


\subsection{Wall Extraction} \label{sec:plane_segmentor}
We use sequential RANSAC to detect and initialize wall planes. In \textit{S-Graphs+}, 
we extract the wall planes from the 3D pointcloud snapshot for a newly registered keyframe, as opposed to our previous work \cite{s_graphs} which extracted wall planes from a continuous stream of 3D pointcloud measurements. This results in efficient detection and mapping of all the wall planes at each keyframe level. Each wall plane extracted at time $t$, $\leftidx{^{L_t}}{\boldsymbol{\pi}}$, is referred to the LiDAR frame $L_t$, we need to convert it to its Closest Point (CP) representation \cite{s_graphs}, and then to the map frame $\leftidx{^{M}}{\boldsymbol{\pi}}$ using the estimated robot pose at time $t$. 
The wall plane normals with their $\leftidx{^M}{{n}_x}$ or $\leftidx{^M}{{n}_y}$ components greater than the $\leftidx{^M}{{n}_z}$ component are classified as vertical planes. Furthermore, normals where $\leftidx{^M}{{n}_x}$ is greater than $\leftidx{^M}{{n}_y}$ are classified as $x$-plane normals, and otherwise they are classified as $y$-plane normals. Finally, planes whose normals' bigger component is $\leftidx{^M}{{n}_z}$ are classified as horizontal planes or ground surfaces.
After initializing each plane in the global map, correspondences are searched for every subsequent plane observation. 
Data association is performed using the Mahalanobis distance between each mapped plane and the newly extracted ones. 

\subsection{Room Segmentation} \label{sec:room_segmentor}

In this work, we present a novel room segmentation strategy capable of segmenting different room configurations in a structured indoor environment improving the room extraction strategy proposed in \cite{s_graphs} which only utilized plane-based heuristics to detect potential room candidates. Proposed room segmentation consists of two steps, \textbf{Free-Space Clustering} and \textbf{Room Extraction}, and the output are the parameters of \textbf{four-wall} and \textbf{two-wall rooms}. 

\textbf{\mbox{Free-Space Clustering.}} Our free-space clustering algorithm divides the free-space graph of a scene into several clusters that should correspond to the rooms of that scene. Given a set of robot poses and a Euclidean Signed Distance Field (ESDF) representation \cite{voxblox} for these poses, we generate a sparse connected graph $\mathcal{G}$ of free spaces using \cite{topological_graphs}. The drift $\leftidx{^M}{\mathbf{x}}_{O}$ estimated after the optimization step in the Back-End (Sec.~\ref{sec:back_end}) is utilized to update the ESDF map also updating the graph $\mathcal{G}$. 
We only maintain an ESDF map and the graph $\mathcal{G}$ up to a certain radius $t_r$ around the robot, clearing the map beyond the radius. 

\begin{figure}[t]
    \centering
    \includegraphics[width=0.35\textwidth]{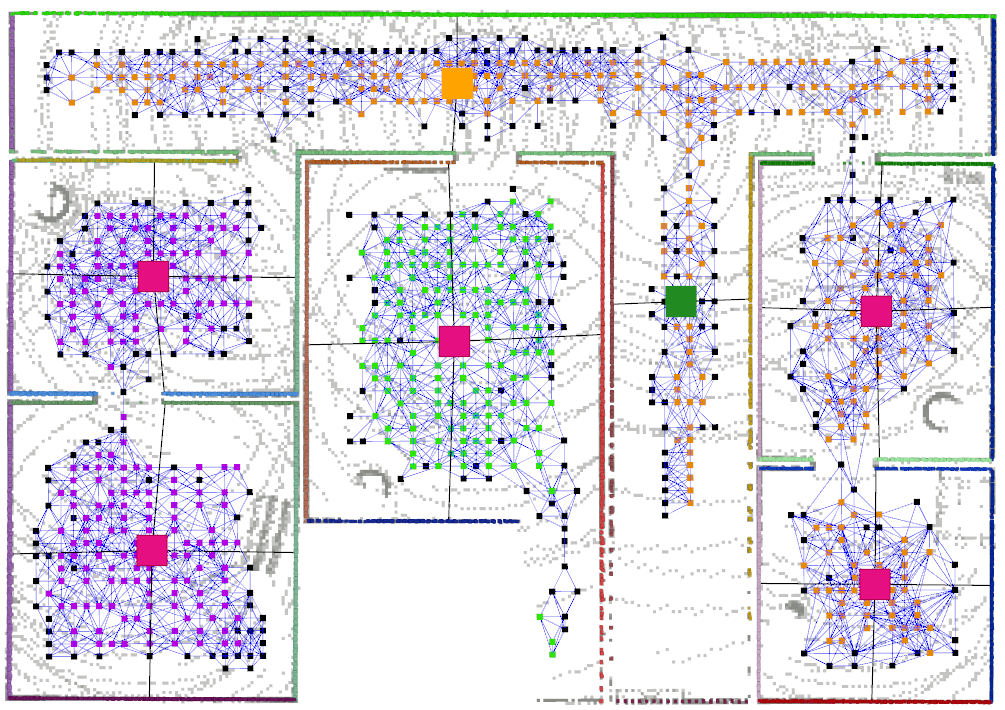}
    \caption{Free space clustering and rooms segmentation, obtained from the estimated wall planes surrounding each cluster. Pink colored squares represent a four-wall room, while yellow and green colored squares represent two-wall rooms in $x$ and $y$ directions respectively. Nodes colored in black are those that are closest to walls and vote for splitting the graph.}
    \label{fig:free_space_generator}
\end{figure}

Given the graph $\mathcal{G}$, we cluster it into different free-space regions as follows. We create a filtered graph $\mathcal{G}_f$ removing the vertices $\boldsymbol{v}_d$ whose distance to obstacles is less than a given threshold $t_\lambda$. We also remove from $\mathcal{G}_f$ all the edges $\boldsymbol{e}_d$ that are connected to the node set $\boldsymbol{v}_d$. We then run the connected components method \cite{graph_theory} on $\mathcal{G}_f$ to divide it into several connected sub-graphs $\mathcal{G}_{f_i}, i \in \{1, \hdots, N\}$. In order to re-connect the deleted vertices $\boldsymbol{v}_d$ and their edges $\boldsymbol{e}_d$ to the filtered sub-graphs $\mathcal{G}_{f_i}$, we check within the entire graph $\mathcal{G}$, each edge $e_{d_i}$ that connects vertex $v_{f_i}$ of a filtered sub-graph $\mathcal{G}_{f_i}$ to the deleted vertex $v_{d_i}$, thus inserting vertex $v_{d_i}$ within $\mathcal{G}_{f_i}$. Using this technique we can obtain disconnected free-space clusters belonging to different rooms, as vertices close to room openings have distances closer to walls (obstacles) and thus vote for disconnecting the graph. Alg.~\ref{alg:free_space_cluster} and Fig.~\ref{fig:free_space_generator} give further details on this free-space clustering. 

\textbf{\mbox{Room Extraction.}} Room extraction uses the free-space clusters $\mathcal{G}_{f_i}$ and the wall planes from a keyframe at time $t$ to detect different room configurations. Wall planes are represented in the map frame as $\leftidx{^M}{\boldsymbol{\Pi}} = [\leftidx{^M}{\boldsymbol{\pi}_{i}}, \ldots, \leftidx{^M}{\boldsymbol{\pi}_{j}}]$, where each plane $\leftidx{^M}{\boldsymbol{\pi}_{i}} = [\leftidx{^M}{\boldsymbol{n}}, \leftidx{^M}{d}]$ is defined by its normal  $\leftidx{^M}{\boldsymbol{n}} = [\leftidx{^M}{n_x}, \leftidx{^M}{n_y}, \leftidx{^M}{n_z}]$ and distance $\leftidx{^M}{d}$ to the origin. 
All extracted wall planes are first categorized as $x$-direction planes $\leftidx{^M}{\boldsymbol{\Pi}_x}$, for which their highest normal component is $n_x$, and $y$-direction planes $\leftidx{^M}{\boldsymbol{\Pi}_y}$ for which the highest normal dimension is $n_y$. $\leftidx{^M}{\boldsymbol{\Pi}_x}$ plane are further classified as $\leftidx{^M}{\boldsymbol{\Pi}_{x_a}}$,  with $n_x>0$, and $\leftidx{^M}{\boldsymbol{\Pi}_{x_b}}$ with $n_x<0$. 
Analogously $\leftidx{^M}{\boldsymbol{\Pi}_{y_a}}$ and $\leftidx{^M}{\boldsymbol{\Pi}_{y_b}}$ represent $y$-planes with positive and negative $n_y$ respectively. 

Given each sub-category of the wall planes, our room extraction method first checks the $L2$ norm between the 3D points of each plane and the vertices of each cluster $\mathcal{G}_{f_i}$, to find the set of walls lying closer to each specific cluster. 

\begin{algorithm}[t]
{\fontsize{6.7}{6.7}\selectfont
\caption{Free-Space Clustering}\label{alg:free_space_cluster}
\textbf{Input}: Free-space graph $\mathcal{G}$, generated using \cite{topological_graphs} \\
\textbf{Output}: Clustered sub-graphs $\mathcal{G}_{f_i}, i \in \{1, \hdots, N \}$ \\
\SetKwFunction{FMain}{visit($v_f$)}
\hrulefill \\
\textit{1: Filter nodes far from obstacles: $\mathcal{G}_{f} \gets \mathcal{G}$ }\\
\hrulefill \\
\For{$v_i \in \boldsymbol{v}$ $\&$ $\boldsymbol{v}$ $\in$ $\mathcal{G}$}{
    \eIf{$v_i.distance < t_\lambda$}{
        $\boldsymbol{v}_d \gets v_{i}$ $\&$ $\boldsymbol{e}_d \gets v_i.edges$\\
    }{
        $\boldsymbol{v}_f \gets v_i$ $\&$ $\boldsymbol{e}_f \gets v_i.edges$ \\
    }
}
$\mathcal{G}_{f} \gets (\boldsymbol{v}_f, \boldsymbol{e}_f)$ \\
\hrulefill \\
\textit{2: Graph clustering by connectivity: $\mathcal{G}_{f_1} \ldots \mathcal{G}_{f_n} \gets \mathcal{G}_f$}\\
\hrulefill \\
$c \gets 0$ \\
\For{$v_{f_i} \in \boldsymbol{v}_f$}{
    $v_{f_i}.visited \gets False$ $\&$ $v_{f_i}.cluster \gets 0$ \\
}
\SetKwProg{Fn}{Function}{:}{\KwRet}
  \Fn{\FMain}{
    \If{$v_{f_i}.visited = False$}{
        $v_{f_i}.visited \gets True$ \\
        \If{$v_{f_i}.cluster = 0$}{
            $c \gets c+1$ \\
            $v_{f_i}.cluster \gets c$ $\&$  $\mathcal{G}_c \gets v_{f_i}$ \\
        }
        \For{$v_{f_n} \in v_{f_i}.neighbours$}{
            $v_{f_n}.cluster \gets c$ $\&$ 
            $visit(v_{f_n})$ \\
        }
    }
  }
\For{$v_{f_i} \in \boldsymbol{v}_f$}{
    $visit(v_{f_i})$
}
\hrulefill \\
\textit{3: Inclusion of deleted vertices and edges in the cluster: $\mathcal{G}_{f_i} \gets (\boldsymbol{v}_d, \boldsymbol{e}_d$)} \\
\hrulefill \\
\For{$v_{d_i} \in \boldsymbol{v}_d$}{
    \For{$v_n \in v_{d_i}.neighbours$}{
        \If{$v_n  \in \mathcal{G}_{f_i}$}{
            $c \gets v_n.cluster$ $\&$ $\mathcal{G}_{f_i} \gets (v_{d_i},e_{d_i}) $
        }
    }
}}
\end{algorithm}
\textbf{\mbox{Four-Wall Rooms.}} 
For a given cluster $\mathcal{G}_{f_i}$, if the room extraction module finds a set of four wall planes 
$\leftidx{^M}{\boldsymbol{\Pi}_s} = [\leftidx{^M}{\boldsymbol{\pi}_{x_{a_1}}}, \leftidx{^M}{\boldsymbol{\pi}_{x_{b_1}}}, \leftidx{^M}{\boldsymbol{\pi}_{y_{a_1}}}, \leftidx{^M}{\boldsymbol{\pi}_{y_{b_2}}}]$ close to the cluster vertices, it is considered as a four-wall room candidate and further tests are carried out. First, the widths $w_x$ and $w_y$ of $\leftidx{^M}{\boldsymbol{\Pi}_{x}} = \{\leftidx{^M}{\boldsymbol{\pi}_{x_{a_1}}}, \leftidx{^M}{\boldsymbol{\pi}_{x_{b_1}}}\}$ and $\leftidx{^M}{\boldsymbol{\Pi}_{y}} = \{ \leftidx{^M}{\boldsymbol{\pi}_{y_{a_1}}}, \leftidx{^M}{\boldsymbol{\pi}_{y_{b_1}}}\}$ should be greater than a given threshold $t_w$, where
\begin{gather} 
    w_x = \big[ \lvert {\leftidx{^M}{d}_{x_{a_1}} \rvert} \cdot \leftidx{^M}{\mathbf{n}}_{x_{a_1}} - {\lvert \leftidx{^M}{d}_{x_{b_1}} \rvert} \cdot \leftidx{^M}{\mathbf{n}}_{x_{b_1}} \big] \nonumber \\ 
    w_y = \big[ \lvert {\leftidx{^M}{d}_{y_{a_1}} \rvert} \cdot \leftidx{^M}{\mathbf{n}}_{y_{a_1}} - {\lvert \leftidx{^M}{d}_{y_{b_1}} \rvert} \cdot \leftidx{^M}{\mathbf{n}}_{y_{b_1}} \big]
    \label{eq:room_width}
\end{gather}

\noindent $\leftidx{^M}d_{x_{a_1}}$ and $\leftidx{^M}d_{x_{b_1}}$ are the plane distances to the origin and $\leftidx{^M}{\mathbf{n}_{x_{a_1}}}$ and $\leftidx{^M}{\mathbf{n}_{x_{b_1}}}$ are the normals of $x$-planes. Similarly, $\leftidx{^M}{d_{y_{a_1}}}$, $\leftidx{^M}{d_{y{b_1}}}$, $\leftidx{^M}{\mathbf{n}_{y_{a_1}}}$ and $\leftidx{^M}{\mathbf{n}_{y_{b_1}}}$ are the distances and normals for $y$-planes. For Eq.~\ref{eq:room_width} to hold true, $\lvert {\leftidx{^M}{d}_{x_{a_1}} \rvert} > \lvert {\leftidx{^M}{d}_{x_{b_1}} \rvert}$ and $\lvert {\leftidx{^M}{d}_{y_{a_1}} \rvert} > \lvert {\leftidx{^M}{d}_{y_{b_1}} \rvert}$. All plane normals are converted to point away from the map $M$ frame as:

\begin{equation} \label{eq:plane_normal_transform}
\leftidx{^M}{\mathbf{n}} = \begin{cases}
     -1 \cdot  \leftidx{^M}{\mathbf{n}} & \text{if $\leftidx{^M}{d} > 0$} \\
     \leftidx{^M}{\mathbf{n}} & \text{otherwise}
    \end{cases}
\end{equation}

If the above test is successful, the 3D points in each wall are checked to be enclosed within the two apposed walls. For example, in-plane points belonging to $\leftidx{^M}{\boldsymbol{\pi}_{x_{a_1}}}$ are checked to lie within the points of $\leftidx{^M}{\boldsymbol{\pi}_{y_{a_1}}}$ and $\leftidx{^M}{\boldsymbol{\pi}_{y_{b_1}}}$. 
Given a room candidate with a planar set $\leftidx{^M}{\boldsymbol{\Pi}_s}$ consisting of four walls, we first calculate the room center as follows:
\begin{gather}
\resizebox{1.\hsize}{!}{$
\leftidx{^M}{\mathbf{r}_{x_i}} = \frac{1}{2} 
\big[ \lvert {\leftidx{^M}{d_{x_{a_1}}} \rvert} \cdot \leftidx{^M}{\mathbf{n}_{x_{a_1}}} - {\lvert \leftidx{^M}{d_{x_{b_1}}} \rvert} \cdot \leftidx{^M}{\mathbf{n}_{x_{b_1}}} \big] + \lvert {\leftidx{^M}{d_{x_{b_1}}} \rvert} \cdot \leftidx{^M}{\mathbf{n}_{x_{b_1}}} \nonumber$}
\\
\resizebox{1.\hsize}{!}{$\leftidx{^M}{\mathbf{r}_{y_i}} = \frac{1}{2} \big[ \lvert {\leftidx{^M}{d_{y_{a_1}}} \rvert} \cdot \leftidx{^M}{\mathbf{n}_{y_{a_1}}} - {\lvert \leftidx{^M}{d_{y_{b_1}}} \rvert} \cdot \leftidx{^M}{\mathbf{n}_{y_{b_1}}} \big] + \lvert {\leftidx{^M}{d_{y_{b_1}}} \rvert} \cdot \leftidx{^M}{\mathbf{n}_{y_{b_1}}} \nonumber$} \\
\leftidx{^M}{\boldsymbol{\rho}_i}  = \leftidx{^M}{\boldsymbol{r}_{x_i}} + \leftidx{^M}{\boldsymbol{r}_{y_i}} \label{eq:finite_room_position}
\end{gather}     

Eq.~\ref{eq:finite_room_position} holds true when \mbox{$\lvert d_{x_1} \rvert > \lvert d_{x_2} \rvert$}. Again, all planes are converted to point away from the origin $M$ using Eq.~\ref{eq:plane_normal_transform}.

Data association for the room node follows two steps. First, the $L2$ norm between the positions of the mapped rooms with the newly detected ones is calculated. Second, the shortlisted rooms using the first step undergo $id$ checks at each wall plane and cases with $id$ mismatch further undergo Mahalanobis distance check.  
This process allows for the identification and merging of duplicate wall planes ($id$ mismatch) for a given room, arising from inaccuracies in the wall plane matching step (Sec.~\ref{sec:plane_segmentor}) as the room and its respective wall plane matching thresholds can be safely tuned to be larger than the single wall plane matching threshold. 

\textbf{Two-Wall Rooms.}
The room extraction method is sometimes able to find only two walls that surround a free-space cluster $\mathcal{G}_{f_i}$. These two-wall rooms can be rooms with some undetected walls or corridor-like structures. If a wall plane set $\leftidx{^M}{\boldsymbol{\Pi}_s} = [\leftidx{^M}{\boldsymbol{\pi}_{x_{a_1}}}, \leftidx{^M}{\boldsymbol{\pi}_{x_{b_1}}}]$ contains two $x$-planes then it is a two-wall room in $x$ direction. Analogously, two-wall rooms in the $y$ direction are composed of opposed $y$-planes. Walls forming two-wall rooms undergo the same checks as four-wall rooms, shown in Eq.~\ref{eq:plane_normal_transform} and Eq.~\ref{eq:room_width}. 
Given the fact that two-wall rooms contain information in either $x$ or $y$ direction, the corresponding center $\leftidx{^M}{\boldsymbol{c}_i}$ of the cluster $\mathcal{G}_{f_i}$ is also utilized to compute the two-wall room center as follows: 
\begin{gather}
\resizebox{1.\hsize}{!}{$\leftidx{^M}{\mathbf{r}_{x_i}} =  \frac{1}{2} 
\big[ \lvert {\leftidx{^M}{d_{x_{a_1}}} \rvert} \cdot \leftidx{^M}{\mathbf{n}_{x_{a_1}}} - {\lvert \leftidx{^M}{d_{x_{b_1}}} \rvert} \cdot \leftidx{^M}{\mathbf{n}_{x_{b_1}}} \big]  + \lvert \leftidx{^M}{{d_{x_{b_1}}} \rvert} \cdot \leftidx{^M}{\mathbf{n}_{x_{b_1}}} \nonumber$} \\
\leftidx{^M}{\boldsymbol{\kappa}_{x_i}} = \leftidx{^M}{\mathbf{r}_{x_i}} +  \big[  \leftidx{^M}{\mathbf{c}_i} - [\ \leftidx{^M}{\mathbf{c}_i} \cdot \leftidx{^M}{\hat{\mathbf{r}}_{x_i}} ] \ \cdot \hat{\leftidx{^M}{\mathbf{r}}_{x_i}} \big]
\label{eq:infinite_room_position}
\end{gather}
\noindent where $\leftidx{^M}{\boldsymbol{\kappa}_{x_i}}$ is the two-wall room center in $x$ direction,  $\leftidx{^M}{\hat{\mathbf{r}}_{x_i}} = \leftidx{^M}{\mathbf{r}_{x_i}} / \| \leftidx{^M}{\mathbf{r}_{x_i}}\|$, and $\leftidx{^M}{\boldsymbol{c}_i}$ is the cluster center obtained from the endpoints of the cluster $\mathcal{G}_{f_i}$ as:
\begin{gather}
\leftidx{^M}{{c}}_{x_i} = \frac{1}{2} \big[ \leftidx{^M}{p}_{x_{1}} - \leftidx{^M}{p}_{x_2} \big] + \leftidx{^M}{p}_{x_2} \nonumber \\  \leftidx{^M}{{c}_{y_i}} = \frac{1}{2} \big[ \leftidx{^M}{p}_{y_1} - \leftidx{^M}{p}_{y_2} \big] + \leftidx{^M}{p}_{y_2} \nonumber \\
\leftidx{^M}{\mathbf{c}_i} = \big[ \leftidx{^M}{c_{x_i}}, \leftidx{^M}{c_{y_i}} \big] 
\end{gather}

\noindent where $\leftidx{^M}{p}_{x_{1}}$ to $\leftidx{^M}{p}_{y_{2}}$ are the cluster endpoints. 
Two-wall room center in $y$ direction can be calculated analogously. 

Data association of two-wall rooms follows a similar concept as four-wall rooms. In the case of a two-wall room in the $x$ direction, we first compute the $L2$ norm along the $x$-axis of the two-wall room center followed by the $id$ check of individual wall planes. Cases with $id$ mismatch further undergo $L2$ norm check of the planar points between the detected and the mapped wall planes. 

Detected four and two-wall rooms are optimized along with their corresponding wall planes in the back-end explained in Sec.~\ref{sec:back_end}. 

\subsection{Floor Segmentation} \label{sec:floor_segmentor}
The floor segmentation module extracts the widest wall planes within the current explored floor level by the robot which can then be used to calculate the center of the current floor level. 
Our floor segmentation utilizes the information from all mapped walls to create a sub-category of wall planes as described in the room segmentation (Sec.~\ref{sec:room_segmentor}) as, $\leftidx{^M}{\boldsymbol{\Pi}_{s_t}}$ where $t= \{1, \hdots, T \}$. After receiving a complete plane set it computes the widths $\boldsymbol{w}_x$ between all $x$-direction planes and similarly $\boldsymbol{w}_y$ for $y$-direction planes using Eq.~\ref{eq:room_width}. The wall plane set with the largest $w_x$ and $w_y$ is the chosen candidate for the current floor level. These planar pairs in both $x$ and $y$ direction undergo an additional dot product check between their corresponding normal orientations, $|\mathbf{n}_{x_{a_1}} \cdot \mathbf{n}_{x_{b_1}}| < t_n$ and $|\mathbf{n}_{y_{a_1}} \cdot \mathbf{n}_{y_{b_1}}| < t_n$, to remove wall planes originating outside the building structure. The floor segmentation computes the floor center node using the obtained wall plane candidates following Eq.~\ref{eq:finite_room_position}.  
Whenever the robot ascends or descends to a different floor level, the newly mapped wall planes are incorporated with the new floor, and the current floor center is computed only using the wall planes at that floor.     


\section{Back-End} \label{sec:back_end}

The back-end is responsible for creating and optimizing the four-layered \textit{S-Graphs+} summing the individual cost functions of each layer, explained in detail as follows.


\textbf{Keyframes.} This layer creates a factor node $\leftidx{^M}{\boldsymbol{x}}_{R_t} \in SE(3)$ with the robot keyframe pose at time $t$ in the map frame $M$. The pose nodes are constrained by pairwise odometry readings between consecutive poses as in \cite{s_graphs}. 

\textbf{Walls.} This layer creates the planar factor nodes for the wall planes extracted by the wall segmentation (Sec.~\ref{sec:plane_segmentor}). The planar nodes are factored as $\leftidx{^M}{\boldsymbol{\pi}} = [\leftidx{^M}\phi, \leftidx{^M}\theta, \leftidx{^M}d]$, where $\leftidx{^M}\phi$ and $\leftidx{^M}\theta$ stand for the azimuth and elevation of the plane in frame $M$. The planar nodes are constrained with their corresponding keyframes using pose-plane constraints as in \cite{s_graphs}.  
The room segmentation module utilizes mapped walls at current keyframe $k_t$ (Sec.~\ref{sec:room_segmentor}) to identify different room candidates, whereas mapped walls from all the mapped keyframes $\boldsymbol{k} = \{k_1, \hdots, k_T\}$ are utilized by the floor segmentation module (Sec.~\ref{sec:floor_segmentor}) to identify the center of the floor level.    

\textbf{Rooms.} 
The rooms layer receives the extracted room candidates and their corresponding wall planes from the room segmentation module (Sec.~\ref{sec:room_segmentor}) to create appropriate constraints between them. 

\textit{\textbf{Four-Wall Rooms}}: 
We propose a novel edge formulation that minimizes in a single cost function the room node (generated from its center) and its four mapped wall planes, as opposed to \cite{s_graphs} which comprised of individual cost functions for room and the wall planes. The proposed cost function can be written as:
\begin{multline} \label{eq:finite_room_node}
    c_{\boldsymbol{\rho}} (\leftidx{^M}{\boldsymbol{\rho}}, \big[ \leftidx{^M}{\boldsymbol{\pi}_{x_{a_i}}}, \leftidx{^M}{\boldsymbol{\pi}_{x_{b_i}}}, \leftidx{^M}{\boldsymbol{\pi}_{y_{a_i}}}, \leftidx{^M}{\boldsymbol{\pi}_{y_{b_i}}}\big]) \\ = \sum_{t=1, i=1}^{T, S} \| \leftidx{^M}{\hat{\boldsymbol{\rho}}_i} - {{f(\leftidx{^M}{\tilde{\boldsymbol{\pi}}_{x_{a_i}}}, \leftidx{^M}{\tilde{\boldsymbol{\pi}}_{x_{b_i}}}, \leftidx{^M}{\tilde{\boldsymbol{\pi}}_{y_{a_i}}}, \leftidx{^M}{\tilde{\boldsymbol{\pi}}_{y_{b_i}}})}} \| ^2_{\mathbf{\Lambda}_{\boldsymbol{\tilde{\boldsymbol{\rho}}}_{i,t}}}
\end{multline}
\noindent Where $\leftidx{^M}{\hat{\boldsymbol{\rho}}_i}$ is the estimated four-wall room center obtained from Sec.~\ref{sec:room_segmentor} and $f(\leftidx{^M}{\tilde{\boldsymbol{\pi}}_{x_{a_i}}}, \leftidx{^M}{\tilde{\boldsymbol{\pi}}_{x_{b_i}}}, \leftidx{^M}{\tilde{\boldsymbol{\pi}}_{y_{a_i}}}, \leftidx{^M}{\tilde{\boldsymbol{\pi}}_{y_{b_i}}})$ is the function mapping the four wall planes estimated to a four-wall room center using 
Eq.~\ref{eq:finite_room_position}. 
Compared to \cite{s_graphs} which only included scalar values $d$ in the room center computation, Eq.~\ref{eq:finite_room_position} now includes both the normal direction $n$ and distance $d$.  
The goal of this cost function is to maintain the structural consistency between the four planes forming the room. 
%

\textit{\textbf{Two-Wall Rooms}}: We propose a similar improved cost function to minimize room nodes and their two corresponding wall planes as follows:
\begin{multline} \label{eq:infinite_room_node}
    c_{\boldsymbol{\kappa}}(\leftidx{^M}{\boldsymbol{\kappa}_i},\big[\leftidx{^M}{\boldsymbol{\pi}_{x_{a_1}}}, \leftidx{^M}{\boldsymbol{\pi}_{x_{b_1}}}, \leftidx{^M}{\textbf{c}_i}\big]) \\ = \sum_{t=1,i=1}^{T,K} \| \leftidx{^M}{\hat{\boldsymbol{\kappa}}_i} - f(\leftidx{^M}{\tilde{\boldsymbol{\pi}}_{x_{a_1}}}, \leftidx{^M}{\tilde{\boldsymbol{\pi}}_{x_{b_1}}}, \leftidx{^M}{\textbf{c}_i}) \| ^2_{\mathbf{\Lambda}_{\boldsymbol{\tilde{\boldsymbol{\kappa}}}_{i,t}}}
\end{multline}
\noindent $\leftidx{^M}{\textbf{c}_i}$ is the cluster center, which is kept constant during the optimization, and $\leftidx{^M}{\hat{\boldsymbol{\kappa}}_i}$ is the estimated two-wall room center in $x$ direction obtained from Sec.~\ref{sec:room_segmentor}.
$f(\leftidx{^M}{\tilde{\boldsymbol{\pi}}_{x_{a_1}}}, \leftidx{^M}{\tilde{\boldsymbol{\pi}}_{x_{b_1}}}, \leftidx{^M}{\textbf{c}_i})$ maps the two wall planes along with its cluster center to a room center using Eq.~\ref{eq:infinite_room_position}. When comparing with \cite{s_graphs}, Eq.~\ref{eq:infinite_room_position} now includes both orientation $n$, distance $d$, and the cluster center. The cost function to minimize two-wall rooms in $y$ direction follows Eq.~\ref{eq:infinite_room_node} for wall planes $(\leftidx{^M}{\boldsymbol{\pi}_{y_{a_j}}}, \leftidx{^M}{\boldsymbol{\pi}_{y_{b_j}}})$ and cluster center $\leftidx{^M}{\textbf{c}_j}$. 
Duplicate wall plane nodes identified during the four-wall or two-wall room segmentation are constrained by a factor minimizing the difference between their respective parameters. 

\textbf{Floors.} 
The floor node consists of the center of the current floor level calculated from the floor segmentation (Sec.~\ref{sec:floor_segmentor}). We add a cost function between the floor node and all the mapped four-wall rooms at that floor level as follows:
\begin{equation} \label{eq:floor_node}
  c_{\xi}(\leftidx{^M}{\boldsymbol{\xi_i}}, \leftidx{^M}{\boldsymbol{\rho}_i}) = \sum_{t=1,i=1,j=1}^{T,F,S}  \| \leftidx{^M}{\hat{\boldsymbol{\delta}}_{{\xi_i},{\rho_j}}} - f(\leftidx{^M}{\boldsymbol{\xi_i}}, \leftidx{^M}{\boldsymbol{\rho}_j}) \| ^2_{\mathbf{\Lambda}_{\boldsymbol{\tilde{\boldsymbol{\xi}}}_{i,t}}}
\end{equation}
\noindent where $\leftidx{^M}{\hat{\boldsymbol{\delta}}_{{\xi_i},{\rho_j}}}$ stands for the relative distance between the floor $i$ with center ${\boldsymbol{\xi_i}}$ and the four-wall room $j$ with center $\boldsymbol{\rho}_j$, and $f(\leftidx{^M}{\boldsymbol{\xi_i}}, \leftidx{^M}{\boldsymbol{\rho}_j})$ maps the relative distance between the centers of floor node and four-wall room node. 
Two-wall room nodes are constrained with the floor node using the same Eq.~\ref{eq:floor_node}. While the robot navigates in the surroundings and discovers new wall planes, the estimate of the floor node might change due to the insertion of such planes into the map. If the current floor center calculated from the new wall planes gets updated beyond a threshold $t_f$, the estimate of the floor node is updated in the graph accordingly along with the relative distances between the floors and all the rooms. 
\section{Experimental Results}

\begin{figure}[!b]
\centering
\begin{subfigure}[t]{.2\textwidth}
\centering
\includegraphics[width=1\textwidth]{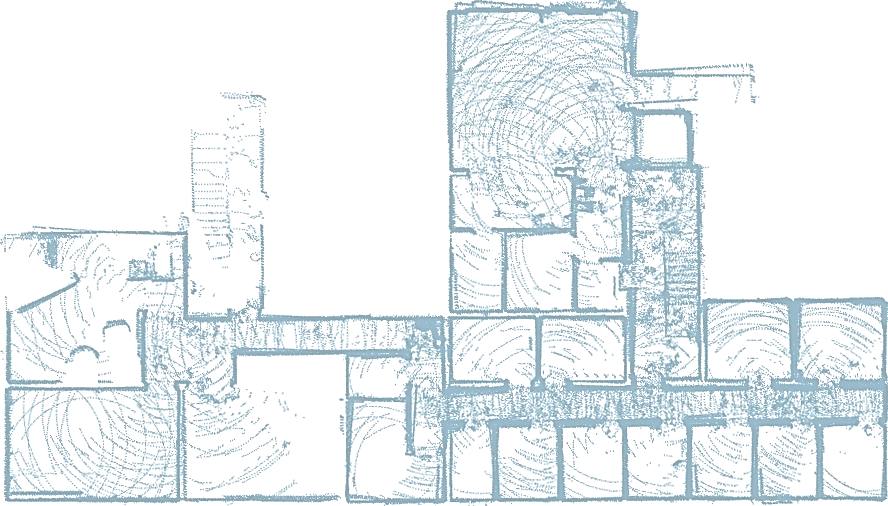}
\caption{{\textit{S-Graphs+}}}
\label{fig:3d_map_s_graphs}
\end{subfigure}
\begin{subfigure}[t]{0.2\textwidth}
\centering
\includegraphics[width=1\textwidth]{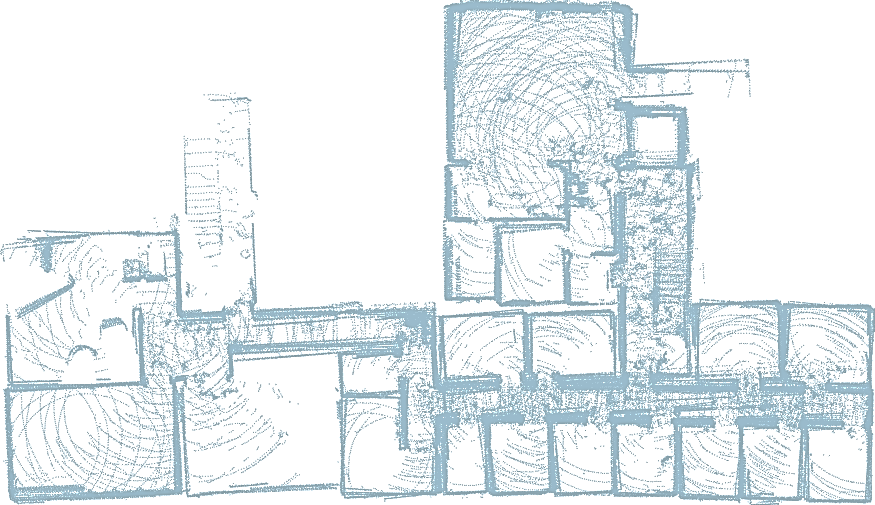}
\caption{HDL-SLAM}
\label{fig:3d_map_hdl_slam}
\end{subfigure}
\begin{subfigure}[t]{0.2\textwidth}
\centering
\includegraphics[width=1\textwidth]{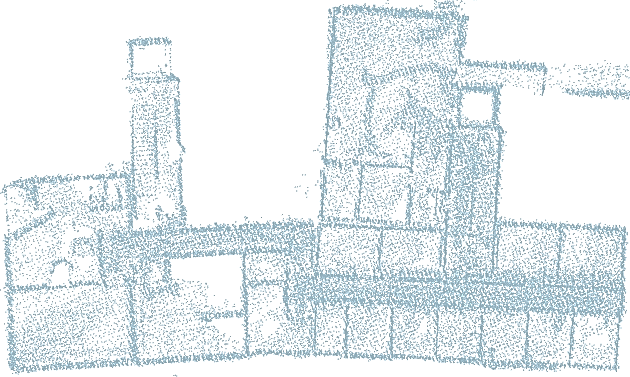}
\caption{ALOAM}
\label{fig:3d_map_aloam}
\end{subfigure}
\begin{subfigure}[t]{0.2\textwidth}
\centering
\includegraphics[width=1\textwidth]{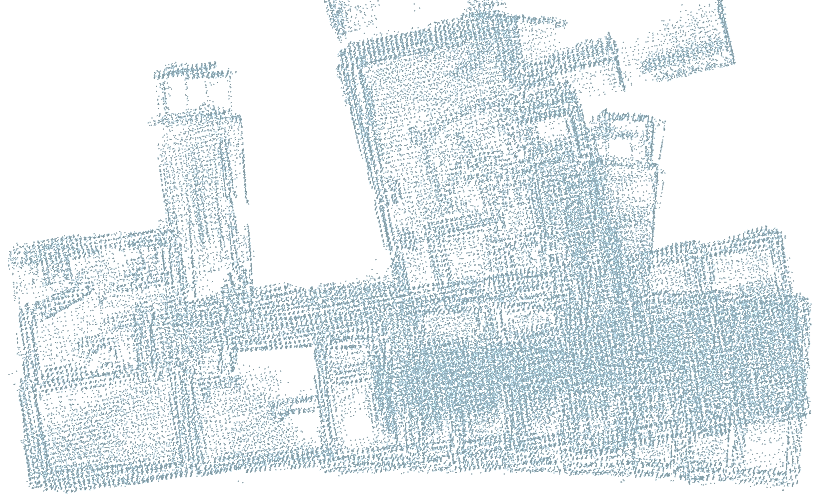}
\caption{FLOAM}
\label{fig:3d_map_floam}
\end{subfigure}
\caption{Maps by \textit{S-Graphs+} and baselines, in-house seq. \textit{C4F0}.}
\label{fig:3d_map_c3f0}
\end{figure}


\subsection{Methodology}
\textit{S-Graphs+} is built on top of its baseline \textit{S-Graphs} \cite{s_graphs} and is validated over several construction sites and office spaces in both simulated and real-world scenarios, comparing it against several state-of-the-art LiDAR SLAM frameworks and its baseline. We utilize VLP-16 LiDAR data in all the datasets. 
To validate the presented novelty, we ablate \textit{S-Graphs+} into \mbox{\textit{S-Graphs+} w. OR} comprising older room detection algorithm from \textit{S-Graphs} but newly proposed room-to-wall plane factors (Sec.~\ref{sec:back_end}) and \mbox{\textit{S-Graphs+} w. OF} with older factors from \textit{S-Graphs} but the newly proposed room detection algorithm (Sec.~\ref{sec:room_segmentor}). 
We do not ablate the proposed floor layer, as currently the floor level (Sec.~\ref{sec:floor_segmentor}) is mostly used to add semantic meaning to the map, without significantly improving the accuracy.  
Furthermore, we compare the room detection of \textit{S-Graphs+} against the heuristics-based one in \textit{S-Graphs}, reporting the precision and recall of the four-walled and two-walled room detections in the real-world scenarios, for which we have the ground truth number of rooms defined in the architectural plans. 

In all the experiments, no fine-tuning of the mentioned thresholds was required and the same prior empirically selected thresholds sufficed for all. 
The ESDF map (Sec.~\ref{sec:room_segmentor}) resolution depends on the LiDAR resolution, which in our case is computed as $0.18$ m vertically and $0.03$ m horizontally, while the map clearing threshold $t_r$ is kept to $10$ m, while $t_{\lambda}$ and $t_{w}$ are kept to $0.8$ m and $0.5$ m respectively. The plane matching threshold (Sec.~\ref{sec:plane_segmentor}) is kept to $0.35$m while the room matching threshold (Sec.~\ref{sec:room_segmentor}) is $1$ m. The dot product threshold $t_n$ between the wall planes (Sec.~\ref{sec:floor_segmentor}) is kept to $0.9$.    

\textbf{Simulated Data.}
We conduct a total of five simulated experiments. \textit{CF1} and \textit{CF2}, are generated from the 3D meshes of two floors of actual architectural plans, while \textit{SE1}, \textit{SE2}, and \textit{SE3}, are performed in additional simulated environments resembling typical indoor environments with different room configurations. We report the ATE against the provided ground truth. Due to absence of odometry from robot encoders, in all simulated experiments the odometry is estimated only from LiDAR measurements. For a fair validation, \textit{S-Graphs+} is run using two different odometry inputs, specifically VGICP \cite{vgicp} and FLOAM \cite{floam}. 
\begin{table}[t]
\setlength{\tabcolsep}{2pt}
\scriptsize
\centering
\caption{Absolute Trajectory Error (ATE) [m], of \textit{S-Graph+} and relevant baselines on simulated data. Best results are boldfaced, second best are underlined. `-' refers to an unsuccessful run.}
\begin{tabular}{l  l | c c c c c | c}
\toprule
\multicolumn{2}{l|}{\textbf{Method}} & \multicolumn{4}{l}{\textbf{Dataset} (m x $10^{-2}$)}  \\
\midrule
{Mapping} & \multicolumn{1}{|l|} {Odometry} & \textit{C1F0} &  \textit{C1F2}  & \textit{SE1} & \textit{SE2} & \textit{SE3} & Avg \\
\midrule
HDL-SLAM \cite{hdl_graph_slam} & \multicolumn{1}{|l|} {VGICP \cite{vgicp}}  & 9.42 &  \underline{2.12} & {2.46} & {10.6} & {6.23} & 6.16 \\ 
ALOAM \cite{loam} & \multicolumn{1}{|l|} {ALOAM} & 9.90 & 8.70 & 15.7 & 40.2 & 19.7 & 18.8 \\
MLOAM \cite{mloam} & \multicolumn{1}{|l|} {MLOAM} & - & 50.2 & 66.1 & -  & 15.7 & 44.0 \\ 
FLOAM \cite{floam} & \multicolumn{1}{|l|} {FLOAM} & 11.7 & 14.5 & 14.6 & 30.5 & 27.6 & 19.8 \\ 
LeGO-LOAM \cite{lego-loam} & \multicolumn{1}{|l|} {LeGO-LOAM}  & - & - & - & - & 74.1 & 74.1 \\
{S-Graphs} \cite{s_graphs} & \multicolumn{1}{|l|}{VGICP} & 5.09 & 2.57 & \underline{2.18}  & 9.10 & \underline{3.86} & 4.56 \\  \midrule
\mbox{\textit{S-Graphs+} w. OR} & \multicolumn{1}{|l|}{VGICP} & \underline{4.95} & {2.52} & \textbf{1.86} & \textbf{8.06} & 4.00 & \underline{4.28} \\
\mbox{\textit{S-Graphs+} w. OF} & \multicolumn{1}{|l|}{VGICP} & 5.31  & 3.00 & \underline{2.18} & \underline{8.40} &  3.93 & 4.56 \\
\midrule
\textit{S-Graphs+ (ours)} & \multicolumn{1}{|l|}{VGICP}  & \textbf{4.47} & \textbf{1.75} & 1.91 & 9.31 & \textbf{3.37} & \textbf{4.16} \\ 
\textit{S-Graphs+ (ours)} & \multicolumn{1}{|l|}{FLOAM} & 5.94 &  11.7 & 5.72 &  9.60 & 19.9 & 10.5 \\ 
\bottomrule
\end{tabular}
\vspace{-4mm}
\label{tab:ate_simulated_data}
\end{table}

\textbf{In-House Dataset.} In all our in-house data we utilize the robot encoders for estimating the odometry. The first two experiments, \textit{C1F1} and \textit{C1F2}, are performed on two floors of a construction site consisting of a single house. 
Additionally, \textit{C2F0}, \textit{C2F1}, and \textit{C2F2} consist of three floors of an ongoing construction site combining four individual houses. \textit{C3F1}, and \textit{C3F2} are two combined houses, while \textit{C4F0} is a basement area with different storage rooms.   
To validate the accuracy of each method in all the real experiments we report the RMSE of the estimated 3D maps against the actual 3D map generated from the architectural plan except for experiment \textit{C4F0}, for which we provide qualitative results due to the absence of a ground truth plan. 


\textbf{TIERS LiDARs dataset.}
We also validate \textit{S-Graphs+} on the public TIERS dataset \cite{tiers_dataset}, recorded by a moving platform in a variety of scenarios. 
Experiments \textit{T6} to \textit{T8} are done in a single small room in which the platform does several passes at increasing speeds. Experiments \textit{T10} and \textit{T11} are performed in a larger indoor hallway with longer trajectories. We report the ATE against the provided ground truth. Due to the absence of encoder readings in this dataset, each baseline method uses its own LiDAR-based odometry. As in the simulated datasets, we validate \textit{S-Graphs+} with VGICP and FLOAM odometry. 

\begin{table}[t]
\setlength{\tabcolsep}{2pt}
\caption{Point cloud RMSE [m] for our in-house real sequences. All methods use odometry from robot encoders. Best results are  boldfaced, second best underlined. `-' refers to an unsuccessful run.}
\scriptsize
\centering
\begin{tabular}{l | c c c c c c c | c}
\toprule
& \multicolumn{3}{l}{\textbf{Dataset} (m x $10^{-2}$)} \\
\toprule
\textbf{Method} & \multicolumn{3}{l}{\textbf{Point Cloud RMSE}} \\
\toprule
 {Mapping}  & \textit{C1F1} &  \textit{C1F2}  & \textit{C2F0} & \textit{C2F1} & \textit{C2F2} & \textit{C3F1} & \textit{C3F2} & Avg \\ 
\midrule
HDL-SLAM \cite{hdl_graph_slam} & 33.5 & 19.8  & 18.5 & 21.1 & 19.5 & 22.9 & 19.4 & 22.1 \\ 
ALOAM \cite{loam} & 52.6 & 33.6 & 34.1  & 45.1 & 29.9 & 36.5 & 43.4 & 39.3  \\ 
MLOAM \cite{mloam} & 45.0 &  27.6 & 40.6 & 32.4 & 23.6  & - & - & 33.8  \\ 
FLOAM \cite{floam}  & 68.5 & 39.2 & 40.2 & 55.5 & 39.5 & 58.3 & 38.8  & 48.6  \\
LeGO-LOAM \cite{lego-loam}  & - & - & 39.2 &  45.5 & - & 52.9 &  50.3 & 47.0 \\
{S-Graphs} \cite{s_graphs} & 33.1 & \textbf{18.9} & 18.4 & 21.8 & \textbf{17.6}  & \underline{22.8} & 22.6 & 22.2 \\ 
\midrule
\mbox{\textit{S-Graphs+} w. OR}  & \textbf{32.4} & \underline{19.0} & 17.7 & \underline{19.9} & \underline{18.2} & 24.1 & 19.4 & 21.5 \\ 
\mbox{\textit{S-Graphs+} w. OF} & \underline{32.8} & \underline{19.0} & \underline{17.0} & 20.1 & \textbf{17.6} & {23.3} & \underline{19.3} & \underline{21.3} \\
\midrule
\textit{S-Graphs+ (ours)} & {32.9} & \textbf{18.9}  & \textbf{16.9} & \textbf{18.9} & \textbf{17.6} & \textbf{22.3} & \textbf{18.7} & \textbf{20.9} \\ 
\bottomrule 
\end{tabular}
\vspace{-2mm}
\label{tab:rmse_real_data}
\end{table}


\subsection{Results and Discussion}

\textbf{Simulated Data.}
Tab.~\ref{tab:ate_simulated_data} showcases the ATE for the simulated experiments.
\mbox{\textit{S-Graphs+} w. OR} results in an average improvement in accuracy of $5.44\%$ over \textit{S-Graphs}, while \mbox{\textit{S-Graphs+} w. OF} shows an average decrease in accuracy of $3.03\%$ over \textit{S-Graphs}. However, the full \textit{S-Graphs+} with both the new room detector and newly proposed factors shows an improved average accuracy of $13.37\%$ over the baseline. 
It can also be seen in Tab.~\ref{tab:ate_simulated_data} \textit{S-Graphs+} is run using two different odometry methods, VGICP and FLOAM, and that it improves the respective odometries by $51.88\%$ and $106.5\%$. 

\begin{figure}[!h]
\centering
\begin{subfigure}[t]{.23\textwidth}
\includegraphics[width=0.97\textwidth]{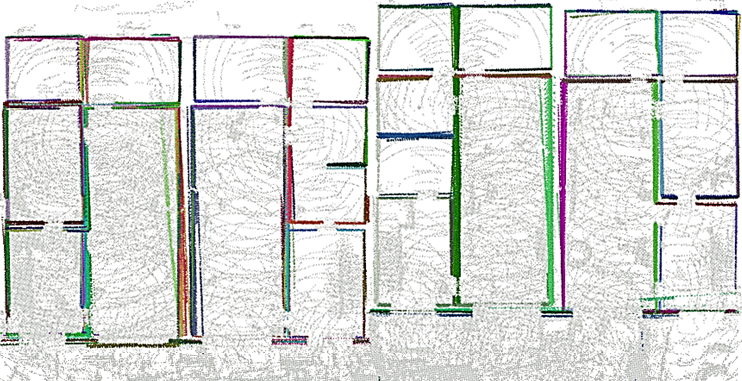}
\caption{{\textit{S-Graphs+}}}
\label{fig:3d_map_s_graphs}
\end{subfigure}
\begin{subfigure}[t]{0.23\textwidth}
\includegraphics[width=0.97\textwidth]{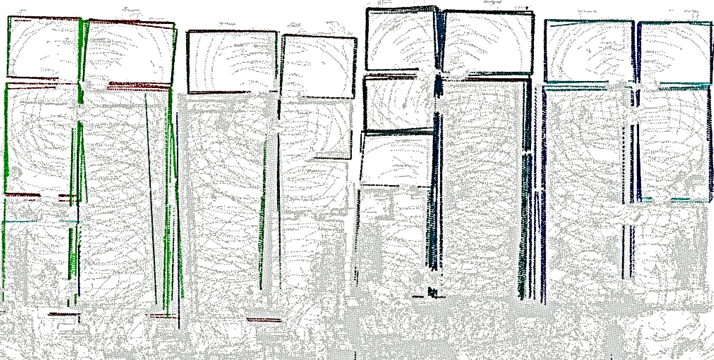}
\caption{\textit{S-Graphs}}
\label{fig:3d_map_aloam}
\end{subfigure}
\caption{\textit{S-Graphs+} and \textit{S-Graphs} maps, in-house seq. \textit{C2F0}.}
\label{fig:3d_map_c2f0}
\vspace{-2mm}
\end{figure}

\textbf{In-House Dataset.} 
Tab.~\ref{tab:rmse_real_data} presents the point cloud RMSE. As it can be observed in the table, \textit{S-Graphs+} outperforms the second-best baseline by a margin of $5.93\%$. \mbox{\textit{S-Graphs+} w. OR} and \mbox{\textit{S-Graphs+} w. OF} individually outperform its baseline by $2.74\%$ and $4.03\%$ respectively. 
For experiment \textit{C4F0}, Fig.~\ref{fig:3d_map_c3f0} shows a top view of the final maps estimated by \textit{S-Graphs+} and three other baselines. Observe the higher degree of accuracy and cleaner map elements in the \textit{S-Graphs+} case, the latest indicating a better alignment for different robot passes. Similarly, observe the precise map generated by \textit{S-Graphs+} in Fig.~\ref{fig:3d_map_c2f0} for experiment \textit{C2F0} when comparing with \textit{S-Graphs}. Fig.~\ref{fig:scene_graph}, shows the entire four-layered \textit{S-Graphs+} for \textit{C2F2} along with its map accuracy. 

\begin{figure}[!b]
    \centering
    \includegraphics[width=0.45\textwidth]{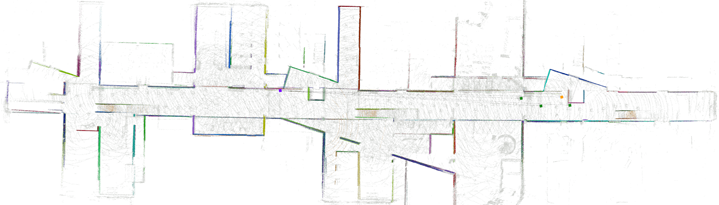}
    \caption{Map estimated by \textit{S-Graphs+} on TIERS sequence \textit{T11}.}
    \label{fig:tiers_dataset}
\end{figure}

\begin{figure}[!b]
\centering
\begin{subfigure}[t]{.2\textwidth}
\centering
\includegraphics[width=0.9\textwidth]{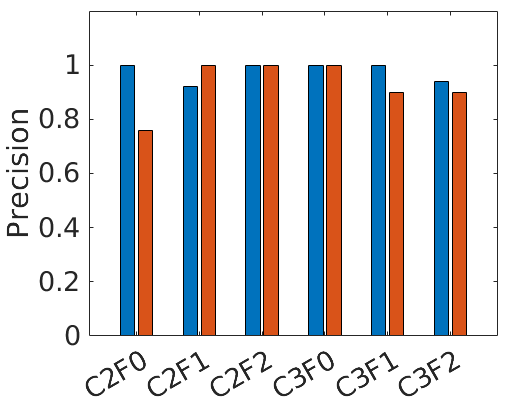}
\label{fig:precision}
\end{subfigure}
~
\begin{subfigure}[t]{0.2\textwidth}
\centering
\includegraphics[width=0.9\textwidth]{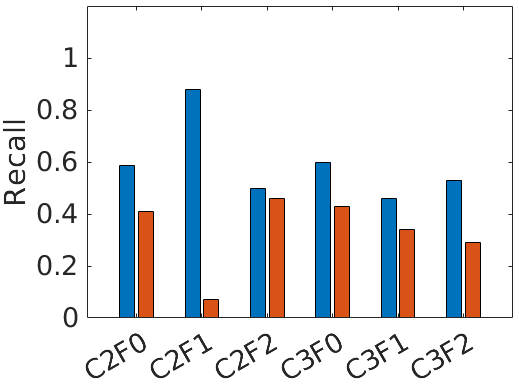}
\label{fig:recall}
\end{subfigure}
\caption{Precision and recall for \textit{S-Graphs+} (blue) and \textit{S-Graphs} (red) on six different scenes of our in-house dataset.}
\label{fig:precision_recall}
\end{figure}

\begin{table}[!t]
\setlength{\tabcolsep}{2pt}
\caption{Computation time [ms] of \textit{S-Graphs+} along the total length of the sequence [s] for In-House dataset.}
\centering
\scriptsize
\begin{tabular}{l | l l l l l l l l}
\toprule
& \multicolumn{3}{l}{\textbf{Dataset}} \\ 
\toprule
 & \multicolumn{6}{l}{\textbf{Computation Time} (mean) [ms]} \\
\toprule
\textbf{Module} & \textit{C1F1} & \textit{C1F2}  & \textit{C2F0} & \textit{C2F1} &\textit{C2F2} & \textit{C3F1} & \textit{C3F2} \\ \midrule
Plane Segmentation & 91.8 & 47.4 & 68.9 & 45.3 & 82.2 & 57.6 & 44.8 \\ 
Room Segmentation & 17.6 & 9.8 & 9.6 & 5.3 & 10.7 & 2.9 & 4.2  \\
Floor Segmentation & 8.1 & 3.4 & 4.6 & 7.2 & 44.7  & 4.0 & 16.9 \\
Back-End & 74.0 & 105.7 & 87.3 & 169.0 & 263.1 & 124.5 & 173.2  \\
\toprule
\textbf{Sequence Length} [s] & 487 & 657 & 238 & 672 & 1044 & 558 & 999 \\
\bottomrule
\end{tabular}
\label{tab:compute_time}
\vspace{-2mm}
\end{table}  

Fig.~\ref{fig:precision_recall} presents the precision/recall of the room detection in \textit{S-Graphs+} and the one in \textit{S-Graphs}. Note how the precision is slightly higher for \textit{S-Graphs+} and, more importantly, the recall is substantially higher for \textit{S-Graphs+}. In particular, the difference is notable for scenarios with complex layouts such as \textit{C2F1} which also improves the final map accuracy (Tab.~\ref{tab:rmse_real_data}). The latest is one of the main strengths of \textit{S-Graphs+}: Extracting a higher number of rooms adds a higher number of constraints leading to more accurate estimates and a better representation. 

Additionally, Tab.~\ref{tab:compute_time} provides a comprehensive overview of the computation time required by each module within \textit{S-Graphs+}. Plane segmentation runtime can vary depending on the area of the wall planes (wall planes with a larger area have a higher number of points, increasing the computation time). The runtime of room segmentation can vary given the number of mapped wall planes in the environment at a given time instant around the robot, while floor segmentation runtime can vary given the current mapped wall planes in the environment (higher number of mapped wall planes increases the computation time).  
The back-end computation time increases with the length of the experiment, as the graph size typically increases with time, but even with sequence lengths of approximately $17$ mins (\textit{C2F2}), all the modules of \textit{S-Graphs+} are able to maintain real-time performance. 

\textbf{TIERS LiDARs dataset.}
Tab.~\ref{tab:ate_tiers_dataset} presents the ATE for all baseline methods and our \textit{S-Graphs+} in the indoor sequences of the public TIERS dataset \cite{tiers_dataset}. On an average \textit{S-Graphs+} with FLOAM odometry gives the best results in all the experiments, improving by $43.2\%$ over FLOAM. Individually \mbox{\textit{S-Graphs+} w. OR} and \mbox{\textit{S-Graphs+} w. OF} improve the average accuracy by $13.36\%$ and $8.32\%$ over \textit{S-Graphs}, while \textit{S-Graphs+} shows average improved accuracy by $14.59\%$ over the second-best baseline.  
Note that all methods perform similarly for small scenes, but differ as scenes become larger. \textit{S-Graphs+} presents significant error reductions for large environments. The strength of our hierarchical representation is particularly evident in scenarios like \textit{T11}, in which \textit{S-Graphs+} utilizing FLOAM odometry increases the FLOAM accuracy by $165.8\%$ (Table~\ref{tab:ate_tiers_dataset} and Fig.~\ref{fig:tiers_dataset}). 
Observe also in  Fig.~\ref{fig:tiers_dataset} the good performance of \textit{S-Graphs+} in non-Manhattan worlds.   
\section{Conclusion}

In this work, we present \textit{S-Graphs+}, a novel four-layered hierarchical factor graph composed of: A \textit{keyframes layer} constraining a sub-set of robot poses at specific distance-time intervals. A \textit{walls layer} constraining the wall plane parameters and linking it to the keyframes. A \textit{rooms layer} modeling detected rooms to their corresponding wall planes and a \textit{floors layer}, denoting the current floor level in the graph and constraining the rooms at that level. To extract this high-level information we also propose a novel room segmentation algorithm using free-space clusters and wall planes and a floor segmentation algorithm extracting the floor centers using all the currently extracted wall planes. We demonstrate an average improvement in the accuracy of $10.67\%$ against the second-best method on our simulated and real experiments covering different indoor environments. 
In future work, we plan to exploit the hierarchical structure of the graph for efficient and faster optimization and validate it over buildings with several floors as well as enhance the reasoning over the graph for improving the detection of different relationship constraints between its semantic elements.  

\begin{table}[t]
\setlength{\tabcolsep}{2pt}
\centering
\caption{Absolute Trajectory Error (ATE) [m], of \textit{S-Graphs+} and relevant baselines on the TIERS dataset \cite{tiers_dataset}. Best results boldfaced, second best underlined.}
\scriptsize
\begin{tabular}{l l | c c c c c | c}
\toprule
\textbf{Method}  & & \multicolumn{4}{l}{\textbf{Dataset} (m x $10^{-2}$)} \\
\toprule
{Mapping} & \multicolumn{1}{|l|} {Odometry}  & \textit{T6} &  \textit{T7}  & \textit{T8} & \textit{T10} & \textit{T11} & Avg \\ 
\midrule
HDL-SLAM \cite{hdl_graph_slam} & \multicolumn{1}{|l|}{VGICP \cite{vgicp}} & 25.6 & 27.3 & \textbf{31.0} & 148.9  & 287.1 & 104.0 \\ 
ALOAM \cite{loam} & \multicolumn{1}{|l|}{ALOAM} & 25.7 & 27.0 & 34.6 & \underline{68.1} & 234.9 & 78.1 \\
MLOAM \cite{mloam} & \multicolumn{1}{|l|}{MLOAM} & 25.7  & \textbf{26.1}  & 33.9 & 263.4 & \textbf{47.4} & 79.3  \\ 
FLOAM \cite{floam} & \multicolumn{1}{|l|}{FLOAM} & 25.8 & \underline{26.3} & 32.4 & 71.3 & 161.1 & 63.4 \\ 
LeGO-LOAM \cite{lego-loam} & \multicolumn{1}{|l|}{LeGO-LOAM} & 27.3 & 33.5 & 36.3 & 140.9 & {68.2} & \underline{61.2}  \\
{S-Graphs} \cite{s_graphs} & \multicolumn{1}{|l|}{VGICP} & 25.6  & 26.8 & 35.1 & 260.1 & 190.1 & 107.5 \\ 
\midrule
\mbox{\textit{S-Graphs+} w. OR} & \multicolumn{1}{|l|}{VGICP} & \underline{25.3} & 26.4 & 32.8 & 133.0 & 173.5 & 78.2 \\
\mbox{\textit{S-Graphs+} w. OF} & \multicolumn{1}{|l|}{VGICP} & 26.5 & 27.1 & 32.9 & 144.9 & 198.3 & 85.9 \\
\midrule
 \textit{S-Graphs+ (ours)}  & \multicolumn{1}{|l|}{VGICP} & 25.6 & 26.6 & 32.9 & 126.6 & {162.3} & 74.8 \\
 \textit{S-Graphs+ (ours)}  & \multicolumn{1}{|l|}{FLOAM} & \textbf{25.2} & {26.5} & \underline{32.1} & \textbf{48.3} & \underline{60.6} & \textbf{38.5} \\
\bottomrule
\end{tabular}
\vspace{-4mm}
\label{tab:ate_tiers_dataset}
\end{table}
\bibliographystyle{IEEEtran}
\bibliography{Bibliography}
\appendix
\section{Appendix}

This section provides all the tables used for computation of the percentage difference between \textit{S-Graphs+} and its relevant baselines. From Table~\ref{tab:percentage_simulated_data_s_graphs} it can be observed \textit{S-Graphs+} \mbox{w. OR} results in an improvement in accuracy of $5.44\%$ over \textit{S-Graphs}, \mbox{\textit{S-Graphs+} w. OF} shows a slight decrease in accuracy of $3.03\%$ over \textit{S-Graphs}, but overall  \textit{S-Graphs+} which combines the new room detector and newly proposed factors shows an improved accuracy of $13.37\%$ over its baseline. 

Table~\ref{tab:percentage_simulated_data_all} presents the percentage improvement of \textit{S-Graphs+} using VGICP odometry over all the baselines in the simulated dataset. \textit{S-Graphs+} shows improvement of $51.88\%$, $411.39\%$, $1298.6\%$, $500\%$, $2098.8\%$ over HDL-SLAM, ALOAM, MLOAM, FLOAM and LeGO-LOAM respectively.

\begin{table}[!htp]
\setlength{\tabcolsep}{4pt}
\scriptsize
\centering
\caption{Percentage increase in accuracy of \textit{S-Graphs+}, \mbox{\textit{S-Graphs+} w. OR} and \mbox{\textit{S-Graphs+} w. OF} with respect to \textit{S-Graphs} on simulated data. Best results are boldfaced.}
\begin{tabular}{l  l | c c c c c | c}
\toprule
\multicolumn{2}{l|}{\textbf{Method}} & \multicolumn{4}{l}{\textbf{Dataset} $\%$ $\boldsymbol{\uparrow}$}  \\
\midrule
{Mapping} & \multicolumn{1}{|l|} {Odometry} & \textit{C1F0} &  \textit{C1F2}  & \textit{SE1} & \textit{SE2} & \textit{SE3} & Avg \\
\midrule
{S-Graphs} \cite{s_graphs} \textit{(Baseline)} & \multicolumn{1}{|l|}{VGICP} & 0 & 0 & 0  & 0 & 0 & 0 \\  
\midrule
\mbox{\textit{S-Graphs+} w. OR} & \multicolumn{1}{|l|}{VGICP} & 2.75 & 1.95 & 14.68  & \textbf{11.43} &  -3.63 & 5.44 \\
\mbox{\textit{S-Graphs+} w. OF} & \multicolumn{1}{|l|}{VGICP} & -4.32  & -16.73 & 0 &  7.69 &  -1.81 & -3.03 \\
\midrule
\textit{S-Graphs+ (ours)} & \multicolumn{1}{|l|}{VGICP}  & \textbf{12.18} & \textbf{31.91} & \textbf{12.39} & -2.31  & \textbf{12.69} & \textbf{13.37} \\ 
\bottomrule
\end{tabular}
\label{tab:percentage_simulated_data_s_graphs}
\end{table}

\begin{table}[!htp]
\setlength{\tabcolsep}{3pt}
\scriptsize
\centering
\caption{Percentage decrease in accuracy of relevant baselines with respect to \textit{S-Graphs+} using VGICP odometry on simulated data. Best results are boldfaced.}
\begin{tabular}{l  l | c c c c c | c}
\toprule
\multicolumn{2}{l|}{\textbf{Method}} & \multicolumn{4}{l}{\textbf{Dataset} $\%$ $\boldsymbol{\downarrow}$}  \\
\midrule
{Mapping} & \multicolumn{1}{|l|} {Odometry} & \textit{C1F0} &  \textit{C1F2}  & \textit{SE1} & \textit{SE2} & \textit{SE3} & Avg \\
\midrule
HDL-SLAM \cite{hdl_graph_slam} & \multicolumn{1}{|l|} {VGICP \cite{vgicp}}  & 110.7 &  {21.14} & 28.80 & 13.86 & 84.87 & 51.87 \\ 
ALOAM \cite{loam} & \multicolumn{1}{|l|} {ALOAM} & 121.5 & 397.1 & 721.9 & 331.8 & 484.6 &  411.4 \\
MLOAM \cite{mloam} & \multicolumn{1}{|l|} {MLOAM} & - & 2768 & 3360 & -  & 364.8 & 1298 \\ 
FLOAM \cite{floam} & \multicolumn{1}{|l|} {FLOAM} & 161.7 & 728.5  & 664.4 & 227.6  & 718.9 & 500 \\ 
LeGO-LOAM \cite{lego-loam} & \multicolumn{1}{|l|} {LeGO-LOAM}  & - & - & - & - & 2099 & 2099 \\
{S-Graphs} \cite{s_graphs} & \multicolumn{1}{|l|}{VGICP} & 13.87 & 46.86 & 14.14  & -2.26 & {14.54} & 17.43 \\  \midrule
\mbox{\textit{S-Graphs+} w. OR} & \multicolumn{1}{|l|}{VGICP} & {10.74} & 44.0 & \textbf{-2.62}  & \textbf{-13.43} &  18.69 & {11.48} \\
\mbox{\textit{S-Graphs+} w. OF} & \multicolumn{1}{|l|}{VGICP} & 18.79  & 71.43 & 14.14 &  -9.77 &  16.62 & 22.24 \\
\midrule
\textit{S-Graphs+ (Baseline)} & \multicolumn{1}{|l|}{VGICP} & \textbf{0} & \textbf{0} & 0 & 0  & \textbf{0} & \textbf{0} \\ 
\textit{S-Graphs+} & \multicolumn{1}{|l|}{FLOAM} & 32.89 &  568.5  & 199.48 &  3.11 &  490.50 & 258.9 \\ 
\bottomrule
\end{tabular}
\label{tab:percentage_simulated_data_all}
\end{table}

\begin{table}[!htp]
\setlength{\tabcolsep}{4pt}
\caption{Percentage increase in accuracy \textit{S-Graphs+}, \mbox{\textit{S-Graphs+} w. OR} and \mbox{\textit{S-Graphs+} w. OF} with respect to \textit{S-Graphs} in real in-house dataset.}
\scriptsize
\centering
\begin{tabular}{l | c c c c c c c | c}
\toprule
& \multicolumn{3}{l}{\textbf{Dataset} (\%) $\boldsymbol{\uparrow}$} \\
\toprule
 {Mapping}  & \textit{C1F1} &  \textit{C1F2}  & \textit{C2F0} & \textit{C2F1} & \textit{C2F2} & \textit{C3F1} & \textit{C3F2} & Avg \\ 
\midrule
{S-Graphs} \cite{s_graphs} & 0 & 0 & 0 & 0 & 0  & 0 &  0 & 0 \\ 
\midrule
\mbox{\textit{S-Graphs+} w. OR}  & \textbf{2.11} & -0.53 & 3.80 & 8.72 & -3.41 & -5.7 & 14.16 &  2.74 \\ 
\mbox{\textit{S-Graphs+} w. OF} & 0.91 & -0.53 & 7.61 & 7.80 & \textbf{0} & -2.19 & 14.60 &  4.03 \\ 
\midrule
\textit{S-Graphs+ (Baseline)} & 0.60 & \textbf{0}  & \textbf{8.15} & \textbf{13.3} & \textbf{0} & \textbf{2.19} & \textbf{17.26} & \textbf{5.93}  \\ 
\bottomrule 
\end{tabular}
\label{tab:rmse_real_data_percentage_s_graphs}
\end{table}

\begin{table}[!htp]
\setlength{\tabcolsep}{3pt}
\caption{Percentage decrease in accuracy of relevant baselines with respect to \textit{S-Graphs+} on real in-house dataset. `-' refers to an unsuccessful run. Best results are boldfaced.}
\scriptsize
\centering
\begin{tabular}{l | c c c c c c c | c}
\toprule
& \multicolumn{3}{l}{\textbf{Dataset} (\%) $\boldsymbol{\downarrow}$} \\
\toprule
 {Mapping}  & \textit{C1F1} &  \textit{C1F2}  & \textit{C2F0} & \textit{C2F1} & \textit{C2F2} & \textit{C3F1} & \textit{C3F2} & Avg \\ 
\midrule
HDL-SLAM \cite{hdl_graph_slam} & 1.82 & 4.76  & 9.47 & 11.64 & 10.80 & 2.69 & 3.74 & 6.42 \\ 
ALOAM \cite{loam} & 59.88 & 77.78 & 101.8  & 138.6 & 69.89 & 63.68 & 132.1 & 91.96 \\ 
MLOAM \cite{mloam} & 36.78 &  46.03 & 140.2 & 71.43 & 34.09  & - & - &  46.94 \\ 
FLOAM \cite{floam}  & 108.2 & 107.4 & 137.7 & 193.7 & 124.43 & 161.4 & 107.5 & 134.4   \\
LeGO-LOAM \cite{lego-loam}  & - & - & 132.0 &  140.7 & - & 137.2 &  169.0 & 82.7 \\
{S-Graphs} \cite{s_graphs} & 0.61 & \textbf{0} & 8.88 & 15.34 & \textbf{0}  & {2.24} &  20.86 & 6.85 \\ 
\midrule
\mbox{\textit{S-Graphs+} w. OR}  & \textbf{-1.52} & {0.53} & 4.73 & {5.29} & 3.41 & 8.07 & 3.74 & 3.46 \\ 
\mbox{\textit{S-Graphs+} w. OF} & -0.3 & 0.53 & {0.59} & 6.35 & \textbf{0} & 4.48 & {3.21} & {2.12}  \\ 
\midrule
\textit{S-Graphs+ (Baseline)} & 0 & \textbf{0}  & \textbf{0} & \textbf{0} & \textbf{0}  & \textbf{0} & \textbf{0} & \textbf{0} \\ 
\bottomrule 
\end{tabular}
\label{tab:rmse_real_data_percentage_all}
\end{table}

Table~\ref{tab:rmse_real_data_percentage_s_graphs} shows the percentage improvement of \mbox{\textit{S-Graphs+} w. OR}, \mbox{\textit{S-Graphs+} w. OF} and \textit{S-Graphs+} over its baseline \textit{S-Graphs} in case of in-house real experiments. \mbox{\textit{S-Graphs+} w. OR} shows an improvement of $2.74\%$ over its baseline, while \mbox{\textit{S-Graphs+} w. OF} shows an improvement of $4.03\%$ with respect to its baseline. \textit{S-Graphs+} combining both the \mbox{\textit{S-Graphs+} w. OR} and \mbox{\textit{S-Graphs+} w. OF} shows an improved performance of $5.93\%$ over its baseline. 

Table~\ref{tab:rmse_real_data_percentage_all} shows the percentage improvement of \textit{S-Graphs+} over its baselines for in-house real experiments. \textit{S-Graphs+} offers improved performance by $6.42\%$, $91.96\%$, $46.94\%$, $134.36\%$, $82.70\%$ over the algorithms HDL-SLAM, ALOAM, MLOAM, FLOAM, and LeGO-LOAM respectively.   

\begin{table}[!htp]
\setlength{\tabcolsep}{4pt}
\centering
\caption{Percentage increase in accuracy of  \textit{S-Graphs+}, \mbox{\textit{S-Graphs+} w. OR} and \mbox{\textit{S-Graphs+} w. OF} with respect to the baseline \textit{S-Graphs} utilizing VGICP odometry. Best results are boldfaced.}
\scriptsize
\begin{tabular}{l l | c c c c c | c}
\toprule
\textbf{Method}  & & \multicolumn{4}{l}{\textbf{Dataset} (\%) $\boldsymbol{\uparrow}$} \\
\toprule
{Mapping} & \multicolumn{1}{|l|} {Odometry}  & \textit{T6} & 
\textit{T7}  & \textit{T8} & \textit{T10} & \textit{T11} & Avg \\ 
\midrule
{S-Graphs} \cite{s_graphs} \textit{(Baseline)} &  \multicolumn{1}{|l|}{VGICP} & 0 & 0 & 0 & 0 & 0 & 0 \\ 
\midrule
\mbox{\textit{S-Graphs+} w. OR} & \multicolumn{1}{|l|}{VGICP} & \textbf{1.17} & \textbf{1.49} & \textbf{6.55} & 48.87 & 8.73 & 13.36 \\
\mbox{\textit{S-Graphs+} w. OF} & \multicolumn{1}{|l|}{VGICP} & -3.52 & -1.12 & 6.27 & 44.29 & -4.31 & 8.32  \\
\midrule
\textit{S-Graphs+} & \multicolumn{1}{|l|}{VGICP} & 0 & 0.75 & 6.27 & \textbf{51.33} & \textbf{14.62} & \textbf{14.59} \\
\bottomrule
\end{tabular}
\label{tab:percentage_ate_tiers_dataset_s_graphs}
\end{table}

\begin{table}[!htp]
\setlength{\tabcolsep}{4pt}
\centering
\caption{Percentage decrease in accuracy of relevant baseline with respect to \textit{S-Graphs+} utilizing FLOAM odometry over the TIERS dataset \cite{tiers_dataset}. Best results are boldfaced.}
\scriptsize
\begin{tabular}{l l | c c c c c | c}
\toprule
\textbf{Method}  & & \multicolumn{4}{l}{\textbf{Dataset} (\%) $\boldsymbol{\downarrow}$} \\
\toprule
{Mapping} & \multicolumn{1}{|l|} {Odometry}  & \textit{T6} &  \textit{T7}  & \textit{T8} & \textit{T10} & \textit{T11} & Avg \\ 
\midrule
HDL-SLAM \cite{hdl_graph_slam} & \multicolumn{1}{|l|}{VGICP \cite{vgicp}} & 1.59 & 3.02 & \textbf{-3.43} & 208.3  & 373.8 & 116.6 \\ 
ALOAM \cite{loam} & \multicolumn{1}{|l|}{ALOAM} & 1.98 & 1.89 & 7.79 & {40.99} & 287.6 & 68.06 \\
MLOAM \cite{mloam} & \multicolumn{1}{|l|}{MLOAM} & 1.98  & \textbf{-1.51}  & 5.61 & 445.3 & \textbf{-21.78} & 85.93 \\ 
FLOAM \cite{floam} & \multicolumn{1}{|l|}{FLOAM} & 2.38 & -0.75 & 0.09 & 47.62 & 165.8 & 43.20 \\ 
LeGO-LOAM \cite{lego-loam} & \multicolumn{1}{|l|}{LeGO-LOAM} & 8.33 & 26.42 & 13.08 & 191.7 &  {12.54} & 50.42 \\
{S-Graphs} \cite{s_graphs} & \multicolumn{1}{|l|}{VGICP} &  1.59 & 1.13 & 9.35 & 438.5 & 213.7 & 132.9 \\ 
\midrule
\mbox{\textit{S-Graphs+} w. OR} & \multicolumn{1}{|l|}{VGICP} & {0.4} & {-0.38} & 2.18 & 175.3 & 186.3 & {72.77} \\
\mbox{\textit{S-Graphs+} w. OF} & \multicolumn{1}{|l|}{VGICP} & 5.16 & 2.26 & 2.49 & 200 & 227.2 & 87.42  \\
\midrule
 \textit{S-Graphs+} & \multicolumn{1}{|l|}{VGICP} & 1.59 & 0.38 & 2.49 & 162.1 & 167.8 & {66.87} \\
 \textit{S-Graphs+ (Baseline)} & \multicolumn{1}{|l|}{FLOAM} & \textbf{0} & 0 & 0 & \textbf{0} & 0 & \textbf{0} \\
\bottomrule
\end{tabular}
\label{tab:percentage_ate_tiers_dataset_all}
\end{table}

Table~\ref{tab:percentage_ate_tiers_dataset_s_graphs} shows the increase in the accuracy of \mbox{\textit{S-Graphs+} w. OR}, \mbox{\textit{S-Graphs+} w. OF} and \textit{S-Graphs+} over its baseline for tiers public dataset. \mbox{\textit{S-Graphs+} w. OR} show improved accuracy of $13.36\%$ and  \mbox{\textit{S-Graphs+} w. OF} shows an improved accuracy of $8.32\%$ over its baseline. While \textit{S-Graphs+} shows an improved accuracy of $14.59\%$ over its baseline.

Table~\ref{tab:percentage_ate_tiers_dataset_all} presents the percentage decrease of the accuracy of baselines over \textit{S-Graphs+} utilizing FLOAM odometry for the tiers public dataset. As can be seen from the table, HDL-SLAM, ALOAM, MLOAM, FLOAM, and LeGO-LOAM show an average decrease in accuracy of $116.64\%$, $68.06\%$, $85.93\%$, $43.20\%$ and $50.42\%$ respectively.

\end{document}